\documentclass[10pt,twocolumn,letterpaper]{article}
\usepackage[utf8]{inputenc}
\usepackage[margin=0.75in]{geometry}
\usepackage{times}
\usepackage{titlesec}
\titleformat{\section}{\normalfont\large\bfseries}{\thesection.}{0.5em}{}
\titleformat{\subsection}{\normalfont\normalsize\bfseries}{\thesubsection}{0.5em}{}
\makeatletter
\renewcommand{\maketitle}{%
  \twocolumn[%
    \begin{center}
    {\Large\bfseries\@title\par}
    \vskip 0.5em
    {\normalsize\@author\par}
    \vskip 1em
    \end{center}
  ]
}
\makeatother
\usepackage{epsfig}
\usepackage{graphicx}
\usepackage{amsmath}
\usepackage{amssymb}
\usepackage{booktabs}
\usepackage{multirow}
\usepackage[breaklinks=true,bookmarks=false]{hyperref}

\begin{document}

\title{SoccerNet 2026 Player-Centric Ball Action Spotting:\\Per-Player Attention with Agreement-Based Ensembling}

\author{Faisal Altawijri, Ismail Mathkour\\TAHAKOM\\{\tt\small faltawijri@tahakom.com, imathkour@tahakom.com}}

\maketitle

\section{System Overview}

Our system follows the two-stage pipeline provided by the organizers~\cite{Ochin2026Footpass}: (1)~a Track-Aware Action Detector (TAAD)~\cite{Singh2023} processes video to produce per-player action logits, and (2)~a Denoising Sequence Transduction (DST)~\cite{Ochin2025DST} autoregressive transformer converts game-state features and TAAD logits into structured event sequences. We contribute improvements to both stages and introduce an agreement-based ensemble strategy. Our final submission achieves \textbf{58.94 Macro-F1} on the challenge set.

\section{Stage 1: TAAD Improvements}

\subsection{Architecture: Temporal Transformer}

We extend the baseline X3D~\cite{FeichtenhoferX3D2020}-based TAAD with a temporal transformer. After ROI-aligned features are extracted per player per frame, we project them to $d{=}256$ dimensions and apply $L{=}4$ transformer encoder layers with 8 attention heads, feedforward dimension 1024, GELU activation, and pre-norm ordering. Learnable positional embeddings (up to 50 frames) provide temporal context. The transformer output is projected to 512 dimensions before the 9-class classification head. This allows the model to reason about temporal patterns across frames rather than classifying each frame independently.

\subsection{Training Modifications}

We identified and fixed a warmup scheduler bug where the learning rate stopped updating prematurely and remained at its warmup value for the remainder of training. Additional training changes: (a)~cosine annealing LR scheduler for all layers except X3D, while X3D uses a fixed LR of $5 \times 10^{-5}$ after warmup; (b)~both learning rates divided by $\sqrt{8}$; (c)~warmup set to 100 steps with no gradient accumulation; (d)~background class weight reduced to 0.05; (e)~tracklet count increased from 4 to 6 at epoch 16. The resulting TAAD logits are used as frozen input to all DST experiments.

\section{Stage 2: DST Improvements}

\subsection{Per-Player Attention}

The baseline encoder treats the 26-role feature vector as a flat input per frame. We introduce a two-stage per-player attention mechanism before the main transformer encoder. Player features (5 game-state channels + 9 TAAD logit channels per role) are reshaped to per-player representations and processed through two stages.

\textbf{Stage 1 (Spatial):} Cross-player spatial attention at each frame. Features are reshaped so that all 26 players at each timestep attend to each other via a transformer encoder layer ($d_p{=}64$, 4 heads), capturing spatial relationships between players at every moment.

\textbf{Stage 2 (Temporal):} Per-player temporal self-attention across frames. Each player's representation from Stage~1 is processed independently across the time dimension with sinusoidal positional encoding, processed by another transformer encoder layer, capturing how individual player states evolve over time.

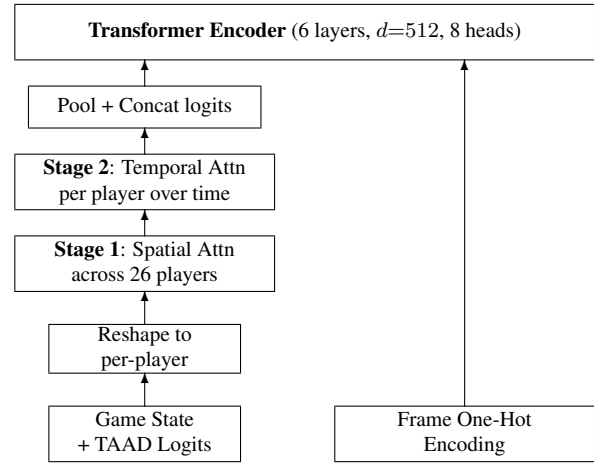
\begin{figure}[h]
\centering
\footnotesize
\setlength{\unitlength}{0.9mm}
\begin{picture}(90,62)
\put(5,0){\framebox(28,8){\parbox{26\unitlength}{\centering Game State\\+ TAAD Logits}}}
\put(19,8){\vector(0,1){5}}
\put(5,13){\framebox(28,7){\parbox{26\unitlength}{\centering Reshape to\\per-player}}}
\put(19,20){\vector(0,1){5}}
\put(0,25){\framebox(38,8){\parbox{36\unitlength}{\centering \textbf{Stage 1}: Spatial Attn\\across 26 players}}}
\put(19,33){\vector(0,1){4}}
\put(0,37){\framebox(38,8){\parbox{36\unitlength}{\centering \textbf{Stage 2}: Temporal Attn\\per player over time}}}
\put(19,45){\vector(0,1){4}}
\put(2,49){\framebox(34,6){\parbox{32\unitlength}{\centering Pool + Concat logits}}}
\put(19,55){\vector(0,1){4}}
\put(47,0){\framebox(38,8){\parbox{36\unitlength}{\centering Frame One-Hot\\Encoding}}}
\put(66,8){\vector(0,1){51}}
\put(0,59){\framebox(85,8){\parbox{83\unitlength}{\centering \textbf{Transformer Encoder} (6 layers, $d{=}512$, 8 heads)}}}
\end{picture}
\vspace{-2mm}
\caption{DST encoder with spatial-first per-player attention. The decoder (not shown) autoregressively generates (frame, action, role) tuples.}
\label{fig:arch}
\vspace{-3mm}
\end{figure}

The attended features are mean-pooled across players and concatenated with the game-state logits per frame, then projected and added to the main encoder input. We found that using only game-state features (positions, velocities, visibility) for the per-player attention, excluding TAAD logits, outperforms using all channels.

\subsection{Spatial-First Attention Order}

We found that processing spatial attention first, then temporal, yields a significant improvement (+1.87\% Macro-F1 on validation, from 0.549 to 0.567). The intuition is that understanding spatial player configurations at each frame provides richer context for subsequent temporal reasoning than the reverse order.

\subsection{Architectural Diversity}

To enable effective ensembling, we train four DST models with different hyperparameters, all using the spatial-first per-player attention mechanism:

\noindent\textbf{Model A} (base): hidden dim 512, 1 player attention layer, $d_p{=}64$ (0.567 val F1)\\
\textbf{Model B}: hidden dim increased to 768 (0.571 val F1)\\
\textbf{Model C}: 2 player attention layers with $d_p{=}128$ (0.572 val F1)\\
\textbf{Model D}: parallel attention, where temporal and spatial branches process the raw input independently and their outputs are summed rather than sequentially (0.569 val F1)

All models use 6 encoder/decoder layers, 8 attention heads, dropout 0.1, learning rate $2.5 \times 10^{-4}$ with exponential decay (factor 0.1 at epochs 3, 6, 8), and are trained for 10 epochs. A checkpoint sweep selects the best-performing epoch per model.

\subsection{Inference: Tau Correction}

During development, we found that selecting the logit adjustment temperature $\tau$ based on per-class optimized Macro-F1 on validation rather than default (raw threshold) Macro-F1 led to suboptimal challenge performance, as per-class thresholds overfit to the 6 validation games (48.0 vs.\ 48.6 on challenge). Using $\tau{=}0$ with default thresholds consistently performs best, and we adopt this for all final predictions.

\section{Ensemble Strategy}

\subsection{Weighted Event Fusion with Agreement Filtering}

Given $N$ models producing event predictions per game, we ensemble using Weighted Event Fusion: (1)~Group all predictions by (team, shirt number, action class). (2)~Cluster predictions within $\pm 12$ frames via greedy assignment to the nearest cluster center, sorted by score descending. (3)~For each cluster with $n$ contributing models: compute the average score across contributing models, multiplied by $(n/N)^{0.5}$ as an agreement boost. (4)~Apply min\_models${=}2$: discard events predicted by fewer than 2 models.

Agreement filtering is the key mechanism: it suppresses false positives (hallucinations from single models) while reinforcing true positives that multiple models agree on. We systematically evaluated NMS, score averaging, max score, and fusion across all model combinations, and fusion with min\_models${=}2$ consistently outperforms all alternatives.

\subsection{Solo Tackle Exception}

Tackle events are extremely rare (26 occurrences across 96 training games) with the lowest per-model F1 ($\sim$0.05--0.15). The agreement filter aggressively removes correct tackle predictions where only one model detects them. We add an exception: tackle predictions bypass the agreement filter and are kept even with single-model support. This preserves recall on the rarest class without affecting other classes.

\section{Results}

\begin{table}[h]
\centering
\small
\begin{tabular}{lcc}
\toprule
Configuration & Val F1 & Challenge F1 \\
\midrule
DST with improved TAAD & 0.541 & 48.6 \\
+ per-player attention & 0.549 & -- \\
+ spatial-first order & 0.567 & 55.1 \\
4-model ensemble & 0.609 & \textbf{58.9} \\
\bottomrule
\end{tabular}
\caption{Progressive improvement. Final ensemble uses Models A--D with fusion, min\_models=2, and solo tackle exception.}
\end{table}

\begin{table}[h]
\centering
\small
\begin{tabular}{lcccccccc}
\toprule
 & dri & pas & crs & thr & sho & hea & tac & blk \\
\midrule
Model A & .803 & .819 & .717 & \underline{.802} & .622 & .403 & .049 & .323 \\
Model B & .804 & .811 & \underline{.749} & .802 & \underline{.630} & .384 & \textbf{.154} & .238 \\
Model C & \underline{.821} & \underline{.824} & .728 & .788 & .629 & .378 & .054 & \underline{.351} \\
Model D & .808 & .819 & .738 & .784 & .634 & \underline{.405} & .100 & .260 \\
\midrule
Ensemble & \textbf{.839} & \textbf{.840} & \textbf{.764} & \textbf{.828} & \textbf{.677} & \textbf{.450} & \underline{.123} & \textbf{.354} \\
\bottomrule
\end{tabular}
\caption{Per-class F1 on validation. \textbf{Bold} = best, \underline{underline} = second best. The ensemble outperforms all individual models on every class except tackle, where Model B is the strongest.}
\end{table}

The ensemble benefits common-to-medium frequency classes (drive through header), where agreement filtering suppresses false positives while preserving recall. The validation-to-challenge gap narrows from $\sim$6 points (single model) to $\sim$2 points (ensemble), indicating improved generalization.

{\small
\bibliographystyle{ieeetr}
\bibliography{egbib}
}

\end{document}